\documentclass[11pt]{article}
\usepackage{coling2020}
\usepackage{times}
\usepackage{url}
\usepackage{latexsym}
\usepackage{subfig}
\usepackage{hyperref}

\newcounter{example}
\newenvironment{example}[1][]{\refstepcounter{example}\par\medskip
   \noindent \textbf{Example~\theexample. #1} \rmfamily}{\medskip}

\title{CEREC: A Corpus for Entity Resolution in Email Conversations}

\author{Parag Pravin Dakle \\
 Department of Computer Science\\
 The University of Texas at Dallas\\
 {\tt paragpravin.dakle@utdallas.edu} \\\And
 Dan I. Moldovan \\
 Department of Computer Science\\
 The University of Texas at Dallas\\
 {\tt moldovan@utdallas.edu} \\}

\date{}

\begin{document}
\maketitle
\begin{abstract}
  We present the first large scale corpus for entity resolution in email conversations (CEREC). The corpus consists of 6001 email threads from the Enron Email Corpus containing 36,448 email messages and 60,383 entity coreference chains. The annotation is carried out as a two-step process with minimal manual effort. Experiments are carried out for evaluating different features and performance of four baselines on the created corpus. For the task of mention identification and coreference resolution, a best performance of 59.2 F1 is reported, highlighting the room for improvement. An in-depth qualitative and quantitative error analysis is presented to understand the limitations of the baselines considered.
\end{abstract}

%
%
\blfootnote{
    %
    %
    %
    %
    \hspace{-0.65cm}  
    This work is licensed under a Creative Commons 
    Attribution 4.0 International License.
    License details:
    \url{http://creativecommons.org/licenses/by/4.0/}.
}

\section{Introduction}
\label{main:introduction}

Entity resolution is defined as linking referring spans of text that point to the same discourse entity by CoNLL 2012 \cite{pradhan-etal-2012-conll} and MUC \cite{grishman-sundheim-1996-message} shared tasks. The corpora used for this task primarily consist of text from news \cite{pradhan-etal-2012-conll,cybulska2014guidelines,recasens2010semeval,grishman-sundheim-1996-message}, web-logs and transcripted dialogs.

This research focusses on the entity resolution task for email conversations. Example \ref{example:task_sample} shows a sample email message and the corresponding entities. The boldfaced tokens represent entities and the numbers beside them represent coreference chain identifiers. An Entity is defined as an object or a group of objects in the real world and a span of text referring to an entity is called a Mention. When all mentions in a text which refer to the same real-world entity are linked together, they form a coreference chain. 

\begin{example} \label{example:task_sample}
    Example of entity resolution task in email conversations \\\\
    \texttt{
Date: Mon, 17 Dec 2001 14:28:03 -0800 (PST)\\
From: \textbf{g..barkowsky@enron.com}\textit{(1)}
To: \textbf{theresa.staab@enron.com}\textit{(2)}\\
Subject: RE: Final Statements and Invoices for November\\
X-From: \textbf{Barkowsky, Gloria G.}\textit{(1)}
X-To: \textbf{Staab, Theresa}\textit{(2)}\\
yes, \textbf{I}\textit{(1)} 'll do this. Do \textbf{you}\textit{(2)} have anything for \textbf{Crestone} \textbf{and Lost Creek}\textit{(3)}?}
\end{example}

Dakle et al. \shortcite{dakle-etal-2020-study} first studied entity resolution in email conversations using a small annotated corpus. Following the same task definition, this paper builds on their work and makes the following key contributions:
\begin{enumerate}
    \item A large corpus for entity resolution in email conversations (CEREC), weakly annotated for mentions and coreference chains, is presented. Detailed corpus statistics are also discussed. The corpus will be released along with the paper\footnote{\url{https://github.com/paragdakle/emailcoref}}.
    \item Experiments with several baseline models are carried out and their results are reported. A qualitative and quantitative error analysis of the results is presented.
\end{enumerate}

The paper is organized as follows: Section 2 reviews related work done on email processing and corpora for emails and entity resolution. Section 3 describes the corpus creation process and reports statistics on the created corpus. It also explores the addition of features and experiments to evaluate the same. Section 4 presents the baseline models, experiments carried out, and the results obtained. Error analysis of the results is covered in Section 5, followed by a discussion on the problem of missing context in Section 6. Section 7 concludes the work done in this paper.

\section{Related Work}
\label{main:related_work}

Email processing has been an active research topic with the earlier works focusing on email classification \cite{cohen1996learning,whittaker1996email,brutlag2000challenges,manco2002towards,klimt2004enron,alkhereyf-rambow-2017-work}. This was later followed by work on intent classification \cite{cohen2004learning}, searching \cite{soboroff2006overview,minkov2008activity}, clustering \cite{huang2008exploring}, and summarization \cite{muresan-etal-2001-combining,lam2002exploiting,newman2003summarizing,nenkova2004facilitating,corston2004task,rambow2004summarizing,carenini2007summarizing,ulrich2008publicly}. 

One of the earliest works highlighting the challenges of the thread-like nature of email conversations was carried out by Lewis and Knowles \shortcite{lewis1997threading}. Murakoshi et. al \shortcite{murakoshi2000construction} proposed the creation of extended contribution trees to better understand the conversation structure of email threads. The impact of coreference resolution on conversations in a threaded format was first studied by Hendrickx and Hoste \shortcite{hendrickx2009coreference} using a corpus of blogs and commented news for opinion mining. Although the impact was negative, it was attributed to the poor performance of the coreference system. Coreference resolution for email conversations is an unexplored problem with, to the best of our knowledge, the only work done by Dakle et. al \shortcite{dakle-etal-2020-study}.

Numerous corpora have been used for email processing over time. University emails \cite{cohen1996learning,cohen2004learning}, email users survey \cite{whittaker1996email,brutlag2000challenges}, private emails \cite{manco2002towards,corston2004task}, simulated emails \cite{lam2002exploiting}, and email archives \cite{nenkova2004facilitating} are few of the initial sources for email corpora. The Enron Email Corpus \cite{klimt2004enron} was the first large public corpus containing emails of 150 employees of the Enron Corporation. Similarly, the Avocado Research Email Collection \cite{oard-2015-avacado} consists of emails from 282 accounts of a now-defunct IT company.

The task of coreference resolution, specifically entity resolution, has received attention in the natural language research community since the 1960s with noun-phrase and pronomial resolution being the early forms of the task. Although multiple corpora released over the years contain a small fraction of telephonic speech text, only a few corpora have focused on the study of the task in a purely conversational setting. Character Identification Corpus \cite{chen2016character} was the first corpus to focus on the entity-linking task in this setting. It was constructed using TV show transcripts with annotations for speakers in a multi-party conversation. Akta{\c{s}} et. al \shortcite{aktacs2018anaphora} used a Twitter corpus to study the performance of Stanford statistical coreference system \cite{clark-manning-2015-entity}. They evaluated a corpus with 185 threads containing 278 coreference chains and reported a mediocre performance by the model.

\section{Corpus}
\label{main:corpus}

\subsection{Seed Corpus}
\label{sub:seed_corpus}

Dakle et al. \shortcite{dakle-etal-2020-study} in their study on entity resolution released a small manually annotated corpus containing 46 email threads from the Enron Email Corpus\footnote{\url{https://www.cs.cmu.edu/~./enron/}} \cite{klimt2004enron}. The Enron Email Corpus is a multi-lingual corpus with the majority of email threads in English. The corpus consists of email threads organized in a directory structure for each user. The annotated corpus consists of 245 email messages with 866 coreference chains containing 5,834 mentions. Each mention refers to an entity of the type PERSON, ORGANIZATION, LOCATION, and DIGITAL\footnote{A digital entity is a media or pointer to a media which is present on some form of digital storage \cite{dakle-etal-2020-study}}. We use the seed corpus (SC) as the starting point\footnote{Only 43 email threads out of the 46 have been used in this work as 3 email threads were discarded due to their overlap with the other email threads in SC} for this work and the underlying Enron Email Corpus as the base corpus to create CEREC. Additionally, the annotation guidelines elaborated by Dakle et al. \shortcite{dakle-etal-2020-study} are followed in this research.

\subsection{Extraction and Filtering}
\label{sub:extraction_and_filtering}

The first step in creating the larger corpus is to shortlist email threads from the Enron Email Corpus. An email thread conversation is considered to be a valid conversation if it contains 4 or more email messages. However, to increase the size of this shortlisted pool of email threads, we do not restrict the scope only to email threads in the \textit{inbox} directory. For each user, email threads in all directories except \textit{all\_documents}, \textit{discussion\_threads}, \textit{drafts}, \textit{deleted\_items}, \textit{sent\_items}, \textit{sent}, \textit{\_sent\_mail}, and \textit{\_sent} are considered. Since, email threads in previous directories are either auto-generated, discarded, or are part of other email threads, they are omitted. A total of 9,724 email threads with a minimum of 4 email messages in each thread are obtained after including additional directories.

On obtaining the initial set of candidate email threads, the following types of email threads are manually filtered from the resulting set:
\begin{enumerate}
    \item Duplicates: An email thread that is part of a larger email thread or is a duplicate belongs to this category. The multi-recipient nature of email conversations results in one email thread possibly being present in directories of multiple users.
    \item No content: Threads in which more than half of the email messages containing no body fall in this category. 
    \item Invalid attachments: The Enron Email Corpus consists of email threads with inline document attachments. Some email threads contain attachments as long hexadecimal strings and hence are labeled as invalid content.
    \item Non-English content: Email threads in the base Enron corpus consists of messages or text in English, Spanish, Russian, German, and French. The scope of this work being restricted to English, email threads containing text in any other language are discarded.
\end{enumerate}

In addition to the above types, we also discard any email threads which overlap partially or fully with those in SC. This is done as eventually SC will be used as the test set for all experiments. After filtering from the initial set, 6144 email threads are obtained.  Table \ref{table:email_filtering_dist} gives a distribution of the initial email threads in each of the filtering categories. Furthermore, the unlabelled corpus contains a total of 37,315 email messages with an average thread length of 6 email messages. 

\begin{table}[h!]
    \centering
    \begin{tabular}{ l  c } 
        \hline
        Email Thread Category & Email Thread Count\\ [0.5ex] 
        \hline
        Duplicates & 2,867\\
        No content & 564\\
        Invalid attachments & 75\\
        Non-English content & 54\\
        Seed corpus overlap & 20\\
        Accepted Email Threads & 6,144\\
        \hline
        Total & 9,724\\
        \hline
    \end{tabular}
    \caption{Distribution of email threads per filtering category}
    \label{table:email_filtering_dist}
\end{table}

\subsection{Annotation}
\label{sub:annotation}

The annotation procedure is divided into two parts: mention annotation and coreference annotation. For both parts, pre-trained SpanBERT \cite{joshi2019spanbert} variant of the model proposed by Joshi et. al \shortcite{joshi2019bert}\footnote{\url{https://github.com/mandarjoshi90/coref}} is used\footnote{Note that here pre-trained SpanBERT implies a pre-trained SpanBERT base model \textbf{not} further trained on the OntoNotes corpus}. Henceforth, we will refer to this model as VanillaSpanBERT (for additional description of the model see \ref{sub:baselines}).

\subsubsection{Mention Annotation}
\label{subsub:mention_annotation}

Given an email thread, correctly identifying spans of text which refer to an entity is the task of mention identification. Here, mention identification task is framed as identifying a single coreference chain which consists of all spans of text referring to a valid entity. A valid entity is an entity of the type PERSON, ORGANIZATION, LOCATION or DIGITAL. Consider Example \ref{example:task_sample}, here the single coreference chain will be \textit{[``g..barkowsky@enron.com", ``theresa.staab@enron.com", ``Barkowsky, Gloria G.", ``Staab, Theresa", ``I", ``you", ``Crestone and Lost Creek"]}. Framing the task in this manner helps in speeding up the annotation process as it eliminates the need to perform architectural changes and carrying out experiments to test each change.

\begin{table}[h!]
    \centering
    \begin{tabular}{ l  c } 
        \hline
        Statistic & Value\\ [0.5ex] 
        \hline
        Added Mentions & 2,106\\
        Corrected Mentions & 344\\
        Deleted Mentions & 325\\
        No-change/Predicted Mentions & 12,056\\
        \hline
        Total Mentions & 13,837\\
        \hline
        Precision & 0.93\\
        Recall & 0.86\\
        F1-score & 0.89\\
        \hline
    \end{tabular}
    \caption{Statistics for changes done during manual correction of predictions obtained on 143 email threads.}
    \label{table:manual_correction_stats}
\end{table}

First, a VanillaSpanBERT model is trained on SC for the mention identification task. Next, this trained model is used to obtain predictions on the unlabelled corpus. From these predictions, approximately 2\% (143 email threads) are manually corrected and a training set of 94 email threads and a validation set of 49 email threads is created. Table \ref{table:manual_correction_stats} shows the count of the type of changes done during the manual correction of these 143 email threads and the corresponding precision, recall, and F1-score of the trained model. In addition to correcting the predictions, we also correct sentence boundaries for these email threads. The remaining 6,001 email threads will be referred to as mention annotated corpus (MAC). The motivation to create a training and validation set is to compare the performance of models trained on gold annotated (94 email threads) and weakly annotated (MAC) training sets, respectively. These models will be referred to as M-VanillaSpanBERT$_{94}$ and M-VanillaSpanBERT$_{6001}$ respectively. Table \ref{table:mention_training_results} reports the results of these two models on SC. From the results, two inferences can be drawn:
\begin{enumerate}
    \item The model M-VanillaSpanBERT$_{6001}$ performs equally well than its counterpart trained on a gold annotated corpus. Weak annotations by definition are either incomplete or contain incorrect annotations. However, based on the correction evaluation statistics (Table \ref{table:manual_correction_stats}) and experiment results, an assumption that they are gold mention annotations for obtaining weak coreference annotation can be made.
    \item The performance of the model M-VanillaSpanBERT$_{6001}$ illustrates the robustness of the model to ignore the noise in the weakly annotated corpus.
\end{enumerate}

Finally, both SC and the training set containing 94 email threads are used to train a VanillaSpanBERT to obtain mention annotations on 6001 email threads, thereby further improving the quality of mention annotations.
\begin{table}[h!]
    \centering
    \begin{tabular}{ l c c c } 
        \hline
        Model & P & R & F1\\ [0.5ex] 
        \hline
        M-VanillaSpanBERT$_{94}$ & 0.94 & 0.82 & 0.8758\\
        M-VanillaSpanBERT$_{6001}$ & 0.95 & 0.808 & 0.8737\\
        \hline
    \end{tabular}
    \caption{Results of two models trained on 94 gold annotated and 6,001 weakly annotated documents respectively.}
    \label{table:mention_training_results}
\end{table}

\subsubsection{Coreference Annotation}
\label{subsub:coref_annotation}

Post completing mention annotation on the unlabelled corpus, the next step is to perform entity coreference annotation. For this task, an approach similar to the one undertaken for obtaining mentions annotations is used. First, a gold validation set is created to assist in understanding the training performance. A set of 34 email threads is selected from the validation set used for mention annotation. Two annotators performed annotation only on the previously gold-annotated mentions. Second, a VanillaSpanBERT model is trained on the coreference annotations of SC to obtain annotations on the MAC. Mention annotations from MAC are provided as input during the coreference annotation process. The final annotated corpus will be referred to as CEREC. Table \ref{table:cerecstats} provides different corpus statistics. Although the corpus contains a large number of mention annotations, 47,013 mentions added during the mention annotation process have not been annotated by the model in this step.

\begin{table}[h!]
    \centering
    \begin{tabular}{ l  c } 
        \hline
        Statistic & Value\\
        \hline
        Number of email threads & 6001 \\
        Number of email messages & 36,448 \\
        Number of words & 6,569,227\\
        Coreference Chains & 60,383 \\
        Annotated Mentions & 445,762 \\
        Annotated Pronouns & 145,615 \\
        Length of longest coreference chain & 489 \\
        Average Length of coreference chains & 7.3822 \\
        \hline
    \end{tabular}
    \caption{CEREC statistics}
    \label{table:cerecstats}
\end{table}



\subsubsection{Environment and Hyperparameters}
\label{subsub:ann_envandpara}

All mention annotation experiments are carried out using the \textit{spanbert\_base} model with a maximum segment length of 256 and on an NVIDIA GeForce GTX 1080 Ti GPU with 8 12gb cores. The base variant of the SpanBERT model trains 2x faster than the large variant only for a loss of 0.1 F1 points. On the other hand, for coreference annotations, \textit{spanbert\_large} with a maximum segment length of 512 outperforms the previous configuration by 7 F1 points. However, this large variant is trained for 10 epochs, and on the CPU due to memory constraints.  The \textit{genre} feature is also removed from all models. All remaining hyperparameters in both settings are left unchanged.

\subsection{Feature Addition}
\label{sub:feature_addition}

Training using additional features like speaker information and genre indicators on top of coreference annotations has proved to be helpful in the past. On the same lines, we evaluate three features specific to conversational texts which have a thread-like structure. 

\begin{enumerate}
    \item Message identifier (MI): For an email thread T containing N email messages, message identifier for a token \textit{x} belonging to message \textit{i} (\textit{i}$\in$\{0, 1, ..., N-1\}) is \textit{i}. 
    \item Section information (SI): An email message is divided into three sections: header, body, and footer\footnote{Footer is defined as the system generated privacy notification or company advertisement. All privacy notifications have been ignored in this work.}. The feature assigns one of the header, body and footer classes to each token in an email message.
    \item Reversing an email (REV): Reversing email messages in a thread refers to ordering the messages as per the time in the email header. This is expected to enhance the model's understanding of the conversation flow in the thread.
\end{enumerate}

For the evaluation, VanillaSpanBERT is used and SC with 43 email threads is used as the training set. The validation set used during the mention annotation process is used with a 14-20 email thread split to create a validation and testing set. A single annotator was used to perform feature annotation on all 77 email threads. Table \ref{table:feature_evaltable} reports results of experiments with permutations of all features using the CoNLL average F1 metric (described in \ref{sub:metrics}). An embedding size of 20 is chosen to encode EI and SI for all feature addition experiments.

\begin{table}[h!]
    \centering
    \begin{tabular}{| l | c |} 
        \hline
        Feature & Avg. F1 (conll) \\
        \hline
        \hline
        VanillaSpanBERT & 55.57\\
        \hline
        \quad + MI & 54.40\\
        \hline
        \quad + SI & \textbf{56.53}\\
        \hline
        \quad + REV & 53.94\\
        \hline
        \quad + REV + MI & 52.15\\
        \hline
        \quad + REV + SI & 54.18\\
        \hline
        \quad + MI + SI & 55.29\\
        \hline
        \quad + REV + MI + SI & 52.94\\
        \hline
    \end{tabular}
    \caption{VanillaSpanBERT evaluation results for all permutations of additional features}
    \label{table:feature_evaltable}
\end{table}

Table \ref{table:feature_evaltable} shows that the addition of SI improves the performance of the model in all scenarios. SI provides information which is useful in identification mentions used for pronoun resolution. All mentions in \textit{To} or \textit{Cc}, or the mention in \textit{From} are used to resolve pronouns like \textit{I}, \textit{you}, \textit{me}, \textit{us}, etc\footnote{This excludes the cases when the sender or an alias of the sender is one of the recipients of the email}.

Reversing the email thread (REV) in temporal order reduces the average F1. This disproves the hypothesis presented before. However, it is important to note that the test size for these experiments consisted of only 20 email threads. Finally, the addition of MI does not help the model. MI provides the model with message boundary information which can be used to merge inter email message clusters but fails to have a positive impact in the current setting.

\section{Experiments}
\label{main:experiments}

\subsection{Baselines}
\label{sub:baselines}

\textbf{Header baseline1 (Hb1):} A simple baseline of resolving pronouns based on the participants in the email header is constructed. All first person singular pronouns (``I", ``me", ``my", ``mine", ``myself") are chained to the sender, and second-person pronouns (``you", ``your", ``yours", ``yourself", ``yourselves") to the recipients respectively. First-person plural pronouns (``we", ``us", ``our", ``ours", ``ourselves") are linked to both the sender and the recipients of the email message. In addition to this, all non-pronomial mentions which are the same or have overlapping words are chained together. This baseline is rule-based and does not consider the surrounding context.

\noindent
\textbf{Header baseline2 (Hb2):} This is similar to Header baseline1 except for how first-person plural pronouns are resolved. In this baseline, all first-person plural pronouns in an email message are chained together into one coreference chain and not to the sender or recipients of that message. Furthermore, each first-person plural pronoun chain in an email message is merged with the corresponding chains in every other message of that email thread.

\noindent
\textbf{c2f-coref (C2F):} The model proposed by Lee et. al \shortcite{lee2018higher} is used for this baseline\footnote{\url{https://github.com/kentonl/e2e-coref}}. This was the first end-to-end neural coreference resolution model. It uses highway LSTMs to generate embeddings for each span and then with a span-ranking model decides which of the previous spans is a suitable antecedent (if any). The inputs to the LSTMs are embedding representations from a language model \cite{peters-etal-2018-deep}.

\noindent
\textbf{VanillaSpanBERT (SBERT):} Joshi et. al \shortcite{joshi2019bert} proposed a BERT \cite{devlin2018bert} version of the C2F model \cite{lee2018higher}. Joshi et. al \shortcite{joshi2019bert} introduced BERT to obtain all input embedding representations. For this baseline, the SpanBERT \cite{joshi2019spanbert} variant of the model is used as the baseline owing to its performance gains.

\subsection{Experimental Setup}
\label{sub:experimental_setup}

The training set for these experiments is CEREC containing 6001 email threads and the validation set contains 34 email threads, the one used for coreference annotation. The SC containing 43 email threads is used as the test set. Mention detection and coreference resolution are the two tasks evaluated in these experiments. The following three experiments are carried out:

\begin{itemize}
    \item Exp1: Use the Hb1 and Hb2 baselines for evaluating coreference resolution given mention annotations as input. Additionally, these baselines also use section information (SI) to identify mentions present in an email header. 
    \item Exp2: Use the C2F and SBERT baselines to evaluate both mention detection and coreference resolution tasks. Compared to the SBERT baselines, the C2F baseline does not enforce a maximum sentence length restriction and has a higher hyperparameter value for maximum training sentences. 
\end{itemize}

The \textit{genre} feature is removed for both C2F and SBERT baselines since it does not apply to this corpus. For the C2F baseline, the hyperparameters \textit{max\_span\_width}, \textit{max\_training\_sentences} and epochs are set to 20, 30 and 10 respectively. This is done to make training tractable on the environment. For the SBERT baseline, the \textit{spanbert\_base} model is used with a maximum segment length of 256, and training is carried out on an NVIDIA GeForce GTX 1080 Ti GPU with 8 12gb cores. 

\subsection{Evaluation Metrics}
\label{sub:metrics}

This work follows the standard experimental setup used in the CoNLL 2012 Shared task. Primary evaluation is done using the unweighted average of MUC, $B^3$, and CEAFE metrics \cite{pradhan-etal-2012-conll}\footnote{\url{https://github.com/conll/reference-coreference-scorers}}. In addition to this, scores using the LEA metric \cite{moosavi-strube-2016-coreference} are also reported.

\subsection{Results}
\label{sub:results}

Table \ref{table:evaltable} shows results of Exp1 and Exp2 for all baselines. First, it can be seen that how first-person plural pronouns are resolved in the header baselines does not have a significant impact on the average F1 score. Second, the average F1 score of SBERT is 0.28 F1 points higher than the C2F baseline. This shows that increasing the maximum sentence length and maximum training sentences does help C2F in outperforming SBERT. Both models perform equally well. Compared to the results reported by Dakle et. al \shortcite{dakle-etal-2020-study}, the SBERT baseline shows an improvement of 5.25 F1 points. Finally, the large difference in F1 scores of the Exp1 baselines and Exp2 baselines is because Exp1 baselines use mention annotations and the SI feature.

\begin{table*}[h!]
    \centering
    \begin{tabular}{| l | c | c | c | c | c | c | c | c | c | c | c | c | p{2em} |} 
        \hline
        & \multicolumn{3}{c}{MUC} & \multicolumn{3}{|c|}{$B^3$} & \multicolumn{3}{|c|}{CEAFE} &
        \multicolumn{3}{|c|}{LEA} &\\
        \hline
        Model & P & R & F1 & P & R & F1 & P & R & F1 & P & R & F1 & Avg. F1 \\
        \hline
        \hline
        Hb1 & 90.2 & 75.1 & \textbf{81.9} & 82 & 65.3 & 72.7 & 61.6 & 74.4 & \textbf{67.4} & 71.1 & 62 & 66.3 & 74 \\
        \hline
        Hb2 & 91.3 & 74 & 81.8 & 87.1 & 64.2 & \textbf{74} & 59.2 & 76.6 & 66.8 & 75.7 & 60.3 & \textbf{67.1} & \textbf{74.2} \\
        \hline
        \hline
        C2F & 86.8 & 64.3 & 73.9 & 72.5 & 46.3 & 56.6 & 67.2 & 35.4 & \textbf{46.3} & 69.4 & 45.5 & 55 & 58.9 \\
        \hline
        SBERT & 87.2 & 64.3 & \textbf{74} & 76.1 & 46.7 & \textbf{57.9} & 63 & 35.9 & 45.7 & 73.9 & 44.6 & \textbf{55.6} & \textbf{59.2} \\
        \hline
    \end{tabular}
    \caption{Evaluation results on SC. Avg. F1 score is computed using MUC, B$^3$ and CEAFE metrics.}
    \label{table:evaltable}
\end{table*}

\section{Error Analysis}
\label{main:error_analysis}

This section presents error analysis performed on the predictions obtained by the baselines on a subset of 15 email threads selected randomly from SC. The selected 15 email threads contain a total of 282 coreference chains with 1261 mentions. To gain an in-depth understanding of the errors, human evaluation is performed. Errors are broadly divided into four categories. These are similar to the categories used by Akta{\c{s}} et. al \shortcite{aktacs2018anaphora} and Dakle et. al \shortcite{dakle-etal-2020-study} in their work. Table \ref{table:errorstats_table} shows the distribution of errors into these categories for each of the baselines.

\subsection{Missing references in the chain}
\label{ref:missing_ref}

A reference that is present in a gold coreference chain but absent in the predicted chains is termed as a missing reference. Hb1 and Hb2 baselines use mention annotations as input to perform coreference chaining. Owing to this reason, only the deep learning baselines are considered for this error category. Missing references are further divided into three types to understand the limitations of the baselines.
\begin{enumerate}
    \item Missing pronoun references: This error type contributes to 8\% of all missing references.
    \item Missing references in email header: A missing email address or name of a participant in the email message present in the email header is considered in this type. This error type contributes to 12\% of all missing references.
    \item Other missing references: All missing non-pronomial references present in the email body are considered in this error type. For C2F and SBERT, the distribution range of these missing references with respect to entity types is: PER - 40\%, ORG - 24\%, LOC - 8-9\%, and DIG - 26-28\%. 
\end{enumerate}

\subsection{Missing chains}
\label{ref:missing_chains}

In this error category, coreference chains that are present in the gold annotations but absent in the predictions are considered. Since Hb1 and Hb2 use mention annotations as input, counts for this error category for these baselines are not reported. The models C2F and SBERT in the original work \cite{lee2018higher,joshi2019bert} were trained on CoNLL 2012 shared task corpus, which did not contain any singletons. Both C2F and SBERT baselines report similar numbers for this error category. About 68-85\% of chains in this error category are of lengths 1 or 2. 

\subsection{Incorrectly chained references}
\label{ref:incorrectly_chained_ref}

All mentions in a coreference chain are considered to refer to the same entity. A mention or reference in a predicted coreference chain which does not refer to the same entity is considered to be incorrectly chained. These references are further broken down into pronoun references and other references. All baselines except C2F report a close count for pronoun references. SpanBERT owing to its higher context capturing capabilities does a better job at resolving pronomial references than C2F. For other references, C2F and SBERT baselines report approximately 1.5 times the counts reported by Hb1 and Hb2. This highlights the effectiveness of rule-based approaches and the possible benefits of having a hybrid approach.

\subsection{Decomposed chains}
\label{ref:decomposed_chains}

A gold coreference chain which is present in the predicted chains in the form of two or more chains is called as a decomposed chain. An email thread consists of multiple email messages. A model may perform well when the scope is restricted to a single email message but may fail to link entity chains belonging to different email messages. In addition to this, paraphrasing of a mention can also result in multiple chains being created. Counts are reported for both the number of original chains and the number of chains that are created. It is evident by the high number of decomposed chains for C2F and SBERT baselines that deep learning models do a worse job of linking chains across email messages and handling paraphrasing.


\begin{table}[h!]
    \centering
    \begin{tabular}{| l | c | c | c | c |} 
        \hline
        Error Category & Hb1 & Hb2 & C2F & SBERT\\
        \hline
        \hline
        Missing references in the chain &  &  &  & \\
        \hline
        \quad Missing pronoun references & - & - & 102 & 101\\
        \hline
        \quad Missing references in email header& - & - & 155 & 160\\
        \hline
        \quad Other missing references & - & - & 224 & 197\\
        \hline
        \hline
        Missing chains & - & - & 131 & 116\\
        \hline
        \hline
        Incorrectly chained references &  &  &  & \\
        \hline
        \quad Pronouns & 60 & 82 & 139 & 98\\
        \hline
        \quad Other & 150 & 137 & 195 & 201\\
        \hline
        \hline
        Decomposed chains &  &  &  & \\
        \hline
        \quad Number of chains decomposed & 50 & 46 & 42 & 63\\
        \hline
        \quad Number of new chains & 134 & 115 & 108 & 156\\
        \hline
    \end{tabular}
    \caption{Error statistics of baselines for different error categories}
    \label{table:errorstats_table}
\end{table}

\section{Problem of missing context}
\label{main:challenges}

Akta{\c{s}} et. al \shortcite{aktacs2018anaphora} and Dakle et. al \shortcite{dakle-etal-2020-study} highlight the challenges encountered for the entity resolution task in a conversational thread-like setting. This section points out an additional challenge corroborates on the difficulty of the task. Conversations using any media generally follow a tree-like structure, where multiple topics may branch off the initial topic but still follow a topic flow. In this flow, every message provides a piece of the whole context which helps in understanding the thread. The deletion of an intermediate message can result in creating ambiguity in the resolution of entities. The deletion of an intermediate email message not only results in the loss of the email text but also the loss of inclusion or exclusion of participants or change of email subject. Carenini et. al \shortcite{carenini-2005-scalable} emphasized this issue in their work on the discovery of hidden emails. 

\section{Conclusion}
\label{main:conclusion}

This paper presents CEREC, the first large annotated corpus for the entity resolution in email conversations task. The corpus consists of 6001 email threads with 60,383 coreference chains. The two steps in the construction of the corpus along with the results of the experiments involved and statistics of the resulting corpus are explained. The construction process is carried out with minimal human intervention. We also evaluate the addition of features specific to text in a conversational thread-like setting. Two rule-based and two deep learning baselines are used for evaluation of the corpus. Qualitative and quantitative error analysis is presented on the predictions obtained using all baselines highlighting the avenues for improvement. Future work will consist of evaluating probable solutions for the entity resolution task. We also plan to conduct additional experiments to understand the effect of features presented in this paper using a larger corpus. 

\bibliographystyle{acl}
\bibliography{anthology,coling2020}

\end{document}